\documentclass[11pt]{article}

\usepackage{graphicx}      
\usepackage{lipsum}        
\usepackage{authblk}       
\usepackage[margin=1in]{geometry}  
\usepackage{booktabs}      
\usepackage[colorlinks=true, citecolor=blue]{hyperref}      
\usepackage{caption}       
\usepackage{booktabs}
\usepackage{multirow}
\usepackage{placeins}
\usepackage{tabularx}
\usepackage{makecell}
\usepackage{hyperref}
\usepackage{adjustbox}
\usepackage{float}         
\usepackage{url}

\renewenvironment{abstract}{%
    \begin{center}%
        \bfseries\abstractname\vspace{-.5em}\vspace{0pt}%
    \end{center}%
    \quotation\noindent%
}{\endquotation}

\newcommand{\keywords}[1]{%
  \begin{quotation} 
  \small\noindent\textbf{Keywords --} #1
  \end{quotation}
}

\title{\textbf{LightPneumoNet: Lightweight Pneumonia Classifier}}

\author[]{Neilansh Chauhan\thanks{Corresponding author: neilanshchauhan4@gmail.com}}
\author[]{Piyush Kumar Gupta}
\author[]{Faraz Doja}

\affil[]{Department of Computer Science, School of Engineering Sciences and Technology, Jamia Hamdard, Hamdard Nagar, New Delhi, 110062, India.
 \protect\\ Email: {neilanshchauhan4@gmail.com, piyush@jamiahamdard.ac.in, farazdoja@jamiahamdard.ac.in}}

\date{} 

\begin{document}

\maketitle 

\begin{abstract}
Effective pneumonia diagnosis is often challenged by the difficulty of deploying large, computationally expensive deep learning models in resource-limited settings. This study introduces LightPneumoNet, an efficient, lightweight convolutional neural network (CNN) built from scratch to provide an accessible and accurate diagnostic solution for pneumonia detection from chest X-rays.
Our model was trained on a public dataset of 5,856 chest X-ray images. Preprocessing included image resizing to 224x224, grayscale conversion, and pixel normalization, with data augmentation (rotation, zoom, shear) to prevent overfitting. The custom architecture features four blocks of stacked convolutional layers and contains only 388,082 trainable parameters, resulting in a minimal 1.48 MB memory footprint.
On the independent test set, our model delivered exceptional performance, achieving an overall accuracy of 0.942, precision of 0.92, and an F1-Score of 0.96. Critically, it obtained a sensitivity (recall) of 0.99, demonstrating a near-perfect ability to identify true pneumonia cases and minimize clinically significant false negatives. Notably, LightPneumoNet achieves this high recall on the same dataset where existing approaches typically require significantly heavier architectures or fail to reach comparable sensitivity levels. The model's efficiency enables deployment on low-cost hardware, making advanced computer-aided diagnosis accessible in underserved clinics and serving as a reliable second-opinion tool to improve patient outcomes.
\end{abstract}



\section{Introduction}
Pneumonia is a serious lung disease characterized by the accumulation of mucus or fluid in the alveoli or bronchioles of the lungs~\cite{Zambare2019}, leading to breathing difficulties. Every year, approximately 450 million individuals are affected by pneumonia worldwide. Coughing, fever, chest pain, fatigue, and loss of appetite are common symptoms of pneumonia~\cite{Grief2018}. While notable advancements in medical technology and healthcare infrastructure have been made, reliable and prompt identification of pneumonia still remains a challenge~\cite{Zar2017}, particularly because of the limitations of existing identification methods~\cite{Varghese2017}. In rural areas, diagnosing pneumonia on time is often difficult due to a lack of healthcare facilities, a shortage of trained medical staff, and limited access to diagnostic imaging equipment~\cite{Sait2019}, which leads to increased morbidity and mortality rates.

In order to overcome the obstacles in quick and precise detection of pneumonia, researchers and healthcare practitioners are employing Artificial Intelligence (AI), particularly advanced deep learning techniques such as Convolutional Neural Networks (CNNs) for automating the process of pneumonia diagnosis~\cite{PVSV2023}. CNN is a Deep Learning algorithm which is inspired by the human visual system to extract meaningful and relevant features from the images~\cite{Ayeni2022}, in our case, medical images (Chest X-Ray images), and can achieve remarkable performances rivaling human experts~\cite{Yadav2019}. 

The performance of CNNs can be further evolved by employing a popular strategy in computer vision called Transfer Learning~\cite{Hashmi2020}. Transfer learning utilizes previously acquired knowledge from one task to enhance performance on a different but related task~\cite{Ali2023}. It allows for the transfer of knowledge from tasks with abundant data, such as general image classification, to tasks with insufficient data~\cite{Pan2010}, such as medical image analysis. This method has shown promising outcomes in improving both the accuracy and sensitivity (recall) of pneumonia detection models, particularly in settings where data scarcity is a challenge~\cite{Alzubaidi2021}. 

It’s important to note that most of the deep learning models proposed for pneumonia diagnosis employ complex and dense architectures. Although these models achieve impressive performance metrics, their complexity introduces several challenges in deployment, practical implementation, training and optimization~\cite{Panwar2017}. As IoT technology continues to grow rapidly, there is an increasing need to deploy these deep learning models on resource-constrained embedded devices, which have constrained memory and computational capabilities~\cite{Hasanpour2016}. So, the deployment of these complex CNN models poses significant practical challenges, including the requirement of substantial computing power during training and inference. In resource-constrained environments, such as rural healthcare facilities, the deployment of such models may be difficult or cost-prohibitive.

In this study, our aim is to develop a lightweight and efficient CNN model, developed from the ground up without relying on popular strategies like transfer learning. The model will have a simpler architecture, making it easy to deploy in environments with limited resources.

\section{Literature Review}
Luka Račić et al. proposed a CNN model for pneumonia detection~\cite{Racic2021}. The dataset employed in this work was introduced by Kermany et al. in 2018~\cite{Kermany2018}, which is publicly accessible on Kaggle. The CNN architecture used in this study comprises 5 convolutional blocks, along with 2 dense layers. However, despite its 8 million parameters, the model achieved a recall of 86.7\%, indicating the ongoing challenge of optimizing both model complexity and performance.

Patrik Szepesi et al. proposed a CNN model~\cite{Szepesi2022} with 10 Convolutional blocks, 7 Dense Layers and a total of 10,604,578 learnable parameters used for pneumonia detection. The dataset employed in this work was introduced by Kermany et al. in 2018~\cite{Kermany2018}. However, despite its 8 million parameters, the model achieved a recall of (97.34±1.56)\%, indicating the ongoing challenge of optimizing both model complexity and performance.

Gaobo Liang et al. proposed a convolutional neural network (CNN) architecture with 49 convolutional layers incorporating a combination of 1×1 and 3×3 filter banks ~\cite{Liang2020}. The proposed architecture is based on a residual network with dilated convolutions and global mean pooling which improves learning in deeper networks. Model performance is improved by using pre-trained weights from ChestX-ray14 dataset (Transfer Learning). However, despite its 49 convolutional layers, the model achieved a sensitivity of 96.7\%, indicating the difficulty in optimizing both model complexity and performance. 

Nada M. et al. presented four distinct models for pneumonia classification, including ResNet152V2, MobileNetV2, CNN and a combination of Long Short-Term Memory (LSTM), and CNN model~\cite{Elshennawy2020}. The dataset utilized in this work was introduced by Kermany et al. in 2018~\cite{Kermany2018}. The ResNet152V2 model with 83,878,529 trainable parameters turned out to be the best model with a recall of 99.43\%, though it comes with the trade-off of a more complex and computationally expensive architecture.

Prateek Chhikara et al. proposed a transfer learning model (InceptionV3) for pneumonia classification~\cite{Chhikara2020}. The dataset employed in this work for model training was introduced by Kermany et al. in 2018~\cite{Kermany2018}. The model utilizes image preprocessing techniques like median filtering, histogram equalization, gamma correction, CLAHE, and JPEG compression enhance feature extraction, along with fine-tuning the InceptionV3 architecture. However, despite using a 316 layered model, the model achieved a recall of 95.7\%, indicating the challenge of optimizing both model complexity and performance.

Ola M. El Zein et al. introduced a hybrid model comprising EfficientNetB0 utilized as a transfer learning-based model alongside support vector machine (SVM) with hinge loss~\cite{ElZein2021}. The pre-trained EfficientNetB0 model serves as feature extractors, followed by an SVM classifier for Pneumonia Classification. The model was trained on Pediatric Pneumonia Chest X-ray~\cite{Kermany2018}, which is a dataset publicly accessible on Kaggle. The hinge loss function-based linear SVM was employed as a replacement for the Sigmoid function within the EfficientNetB0 model. The model attained an accuracy of 97.0\%, with a sensitivity of 95.8\%, indicating the difficulty in optimizing both model complexity and performance. 

Sagar Kora Venu et al. introduced a transfer learning model to lower the training time and minimize generalization error in neural networks~\cite{Venu2020}. Several state-of-the-art AI algorithms, including InceptionResNet, MobileNetV2, Xception, DenseNet201, and ResNet152V2, were trained and fine-tuned for optimal pneumonia classification. A weighted average ensemble of these models was subsequently created to leverage their collective strengths. The model exhibited exceptional performance, with a recall of 99.53\%, though it comes with the trade-off of a more complex and computationally expensive architecture.

Tawsifur Rahman et al. introduced the utilization of four distinct pre-trained deep CNN architectures: AlexNet, ResNet18, DenseNet201, and SqueezeNet, for transfer learning~\cite{Rahman2020}. The dataset employed in this work was introduced by Kermany et al. in 2018~\cite{Kermany2018}. The DenseNet201 model (with 201 Layers) turned out to be the best model with a sensitivity of 99\%, though it comes with the trade-off of a more complex and computationally expensive architecture.

Rohit Kundu et al. proposed an ensemble consisting of three convolutional neural network (CNN) models: GoogLeNet, ResNet-18, and DenseNet-121~\cite{Kundu2021}. For integrating the predictions from these models, a weighted average probability ensemble approach was employed, which was a unique approach. Assessment of the proposed methodology was carried out on two pneumonia chest X-ray datasets, obtained from Kermany~\cite{Kermany2018} and the Radiological Society of North America (RSNA)~\cite{Wang2017}. The ensemble model obtained a recall of 98.80\%, on the Kermany dataset, and 87.02\% on the RSNA challenge dataset, though this lower recall on the RSNA dataset highlights a limitation given the model's complexity.

Ebru Erdem et al. proposed a CNN with six convolutional and six separable convolutional layers, followed by four dense layers (23.9M parameters)~\cite{Erdem2021}. The dataset employed in this work was put forward by Kermany et al. in 2018~\cite{Kermany2018}. Despite 23.9 million parameters, the model, processing images via depth and point convolution, achieved 97.3\% sensitivity, highlighting the challenge of optimizing complexity and performance.

Harsh Bhatt et al. developed a CNN for pneumonia detection using a weighted ensemble of models with kernel sizes (3×3, 5×5, and 7×7)~\cite{Bhatt2023}. Based on Kermany et al.’s dataset~\cite{Kermany2018}, images were converted to grayscale, resized to 180×180 pixels, and combined probabilistically. The ensemble achieved a recall of 99.23\% but lower precision (80.04\%) and accuracy (84.12\%), indicating false-positive concerns, which could reduce reliability in critical applications.

Orlando Iparraguirre-Villanueva et al. proposed four distinct transfer learning-based deep learning models: VGG16, VGG19, ResNet50, and InceptionV3, to classify pneumonia based on chest X-ray imagery~\cite{Iparraguirre2022}. In order to enhance the model’s performance, fine-tuning was employed to modify and adjust the output classifications to align with the specific problem at hand. The ResNet50 model, while achieving a recall of 95.3\% and an accuracy of 68.6\%, exhibits lower classification metrics relative to the model's complexity.

Marwa M. Eid et al. proposed a hybrid method combining a ResNet for feature extraction with an AdaBoost-SVM classifier~\cite{Eid2021}. Using the Kermany et al. dataset~\cite{Kermany2018} their model achieved a high accuracy of 98.13\% and perfect precision of 100\%. However, the recall of 96.40\% suggests a trade-off where the model excels at confirming positive cases at the risk of missing some, a key consideration for clinical screening applications.

Zhongliang Li et al. proposed PNet, a CNN with approximately 1.7 million parameters designed to operate on various image sizes~\cite{Li2019}. When trained on 512x512 images, the model achieved a notable accuracy of 92.79\%, a recall of 92.59\%, and a precision of 89.68\%.These results are relatively low considering the model's complexity, a significant concern in medical applications where high performance is critical.

T. Rajasenbagam et al. developed a pneumonia detection model using a modified VGG19Net architecture and a DCGAN for data augmentation~\cite{Rajasenbagam2021}. The model achieved exceptional performance, including 99.3\% accuracy and 100\% precision. The architecture's complexity, with over 54 million trainable parameters, underscores the trade-offs between achieving high performance and the computational resources required, which can be a consideration for deployment in resource-constrained environments.\\[1em]

\section{Materials and methods}


\subsection{Dataset}
The dataset employed in this work was introduced by Kermany et al. in 2018~\cite{Kermany2018}, which is publicly accessible on Kaggle. The dataset is divided into three folders - train (5216 images), test (624 images) and val (16 images), each having two subfolders - Normal (1583 images) and Pneumonia (4280 images). A total of 5,856 Chest X-ray (JPEG) images were included in the dataset.

\begin{figure}[h!]
    \centering
    \includegraphics[width=\textwidth]{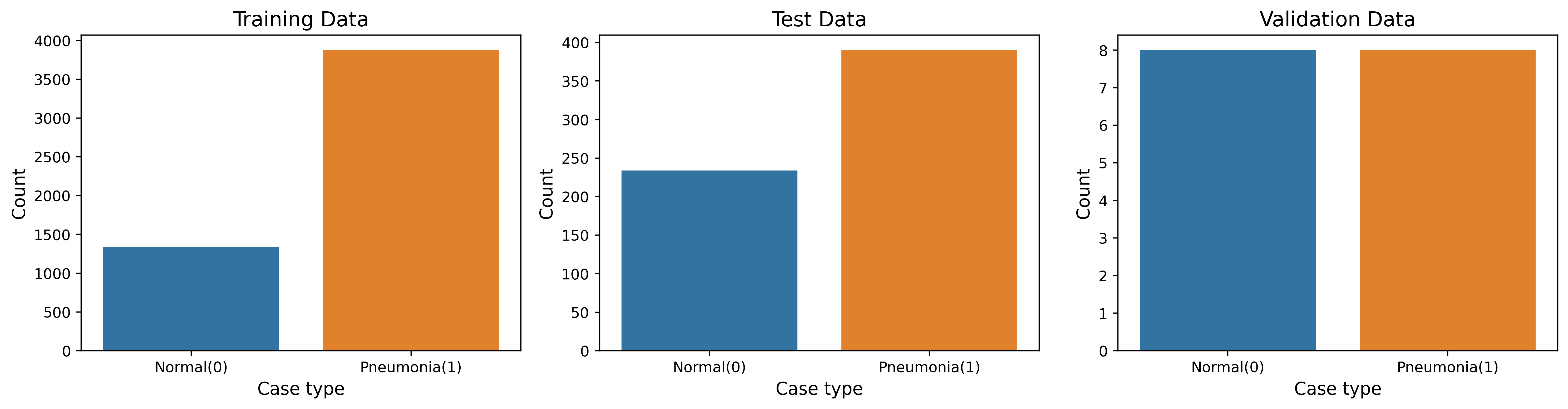} 
    \caption{Class Distribution of Chest X-Ray Images in the Dataset}
    \label{fig:fig1}
\end{figure}

\begin{figure}[h!]
    \centering
    \includegraphics[width=\textwidth]{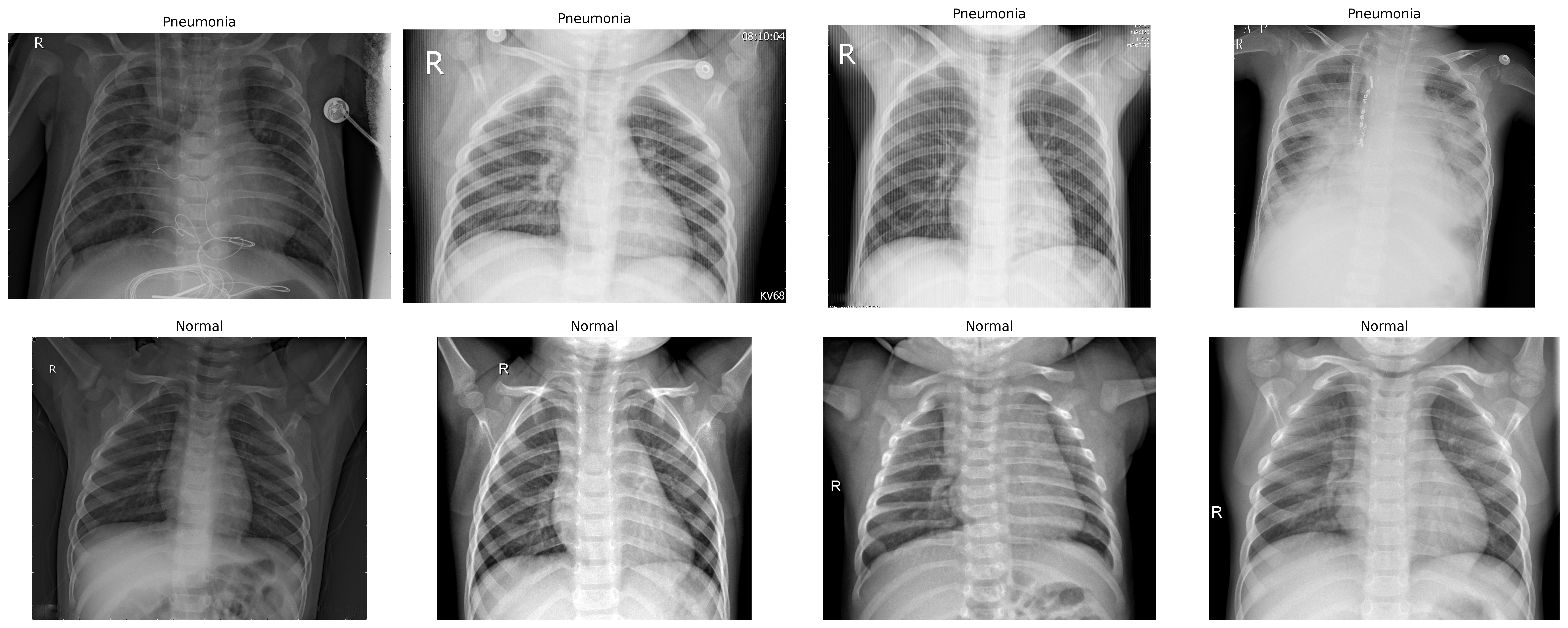} 
    \caption{Chest X-Ray Image Sample}
    \label{fig:fig2}
\end{figure}

\subsection{Preprocessing and Augmentation}

The preprocessing pipeline includes multiple steps before feeding the input images to the CNN. Initially, the images were resized to 224x224 pixels to ensure uniform input dimensions across the dataset. Following this, the RGB images were converted to Grayscale, which surprisingly improved the model’s performance and classification metrics. The pixels were then normalized to a range of [0,1] which enhances the model's training efficiency. Finally, the images were reshaped to a single channel dimension, to conform to the input structure expected by the CNN architecture.

The data augmentation process applies a series of transformations on the Chest X-Ray input images to enhance the diversity of the training data and improve the model’s generalization capabilities, thereby reducing overfitting. The transformations include random rotations of up to 12 degrees, randomly zoom in or out by a factor of ±15\%, and horizontal and vertical shifts within a 15\% range. Additionally, the images are subjected to shear transformations (which distorts the image) of up to 15\%. The  `fill\_mode` is set to 'nearest', ensuring that any newly introduced pixels during these transformations are filled appropriately based on the nearest neighboring pixels.  

\begin{table}[h!]
    \centering
    \caption{Image Preprocessing and Augmentation Settings.}
    \label{tab:preprocessing}
    \begin{tabular*}{0.65\columnwidth}{@{\extracolsep{\fill}}ll} 
        \toprule
        \textbf{Technique} & \textbf{Setting} \\
        \midrule
        Rescale & 1/255 \\
        Rotation Range & 12 \\
        Zoom Range & 0.15 \\
        Width Shift & 0.15 \\
        Height Shift & 0.15 \\
        Shear Range & 0.15 \\
        Fill Mode & Nearest \\
        Horizontal Flip & False \\
        \bottomrule
    \end{tabular*}
\end{table}

\subsection{Model Architecture}
The CNN model consists of 4 convolutional blocks, each with two Convolutional Layers followed by a MaxPooling Layer. The convolutional layers use filters ranging from 16 to 128 to extract features from the input chest X-Ray images, which are grayscale with a size of 224x224.

The first convolutional block uses two convolutional layers with 16 filters and a 5x5 kernel, with a pool size of 3x3 for down-sampling. The second convolutional block increases the filter count to 32, with the same kernel size and pooling as the first block. The third convolutional block increases the filter count to 64, with smaller 3x3 kernels and a pool size of 2x2. The fourth convolutional block uses 128 filters and a 3x3 kernel, with a pool size of 2x2. All the convolutional layers in the model use a ‘valid’ padding, meaning that no padding was applied to the input.

After the convolutional blocks, the output is flattened into a 1D vector, which is followed by two Dense layers, one with 128 neurons and a ReLU activation function. A Dropout layer is added with a rate of 0.2 to reduce overfitting. The final Dense layer has 2 neurons with a softmax activation function, which produces the output probabilities for the two classes.

The key advantage of this CNN model is that it has a total of only 388,082 trainable parameters, requiring just 1.48 MB of memory. This makes the model highly efficient and lightweight, while still maintaining its capacity for effective learning. 

\begin{figure}[H]
    \centering
    \includegraphics[width=\textwidth]{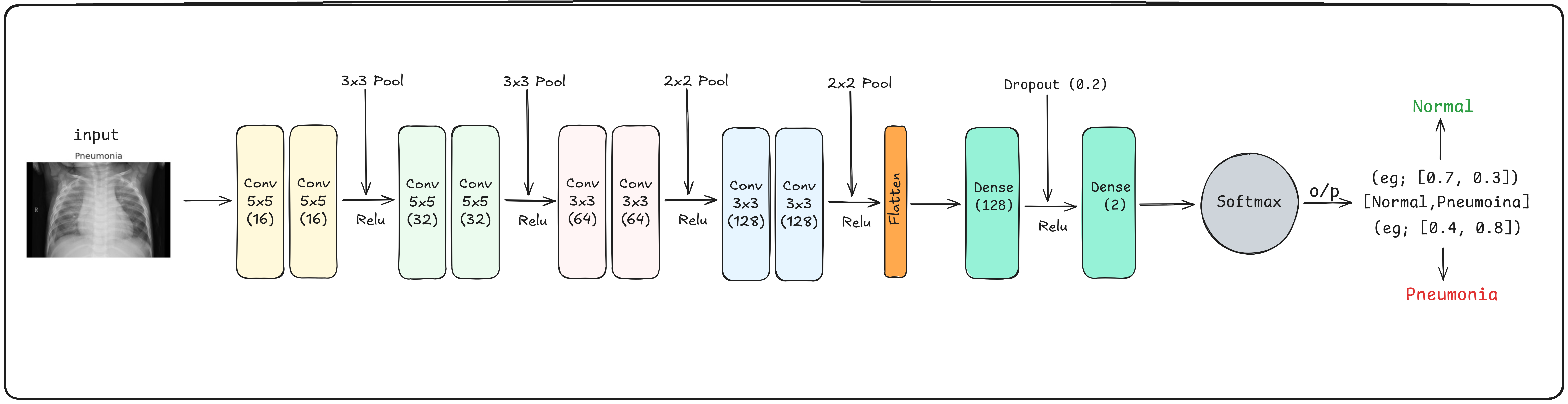} 
    \caption{Model Architecture}
    \label{fig:fig3}
\end{figure}

\begin{figure}[H]
    \centering
    \includegraphics[width=\textwidth]{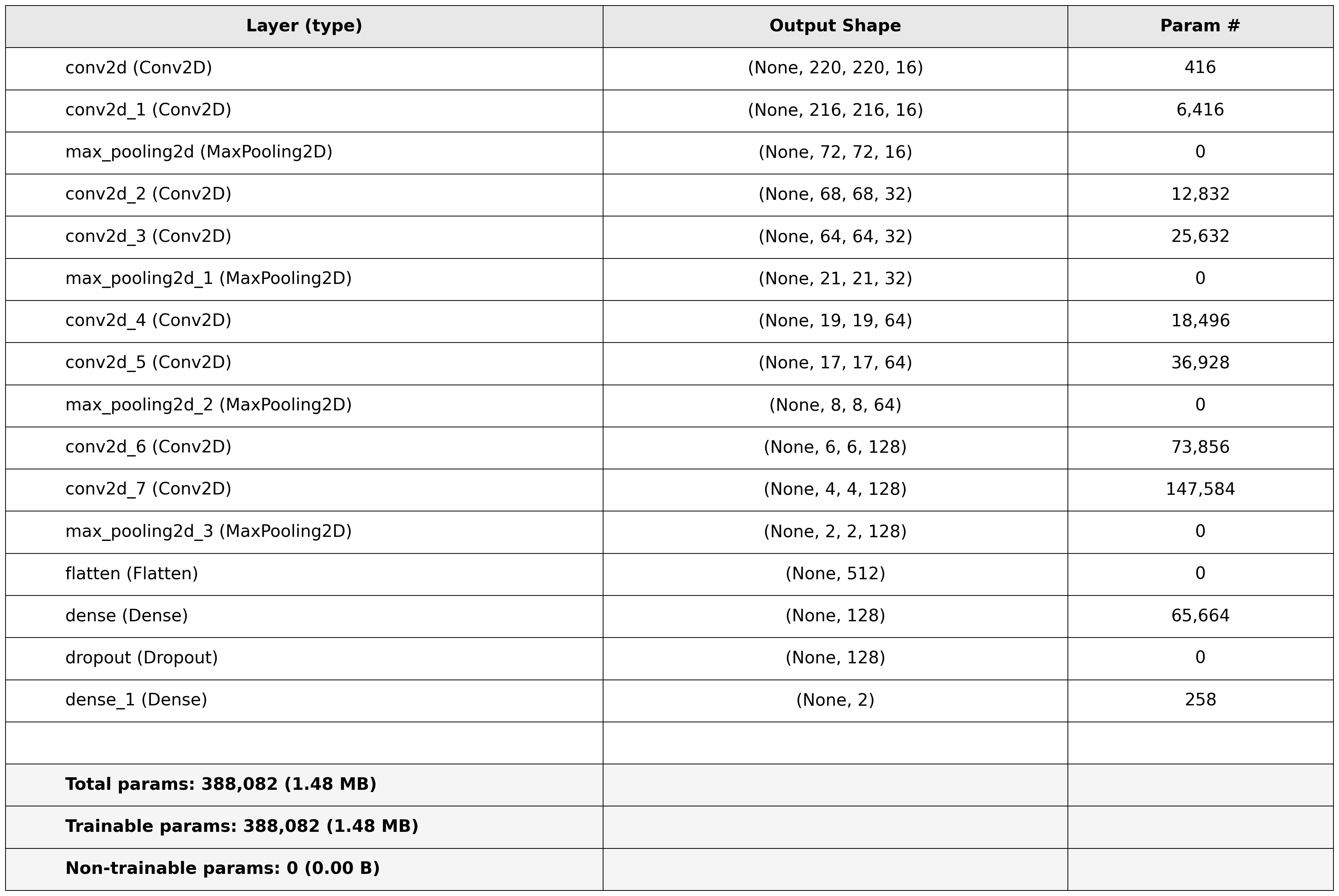} 
    \caption{Model Architecture with Output Shapes and Parameter Counts}
    \label{fig:fig4}
\end{figure}

\FloatBarrier 

\subsection{Hyperparameters}
The Adam optimization algorithm was employed for model compilation, with a learning rate configured at 0.0001 and weight decay of 1e-5. The loss function used is categorical cross-entropy, which is typically used for multi-class classification. We employed categorical cross-entropy with softmax activation as this combination consistently outperformed other loss functions (including binary cross-entropy with sigmoid activation) during preliminary experiments. 

\subsection{Training}
The model is trained for up to 100 epochs, with a batch size of 4. Early Stopping is used to monitor the training loss and stop training if the loss does not reduce after 5 epochs to prevent overfitting. 

A class\_weight parameter is used to handle the class imbalance in our dataset, assigning a weight of 2.0 to class\_0 (Normal) and 1.2 to class\_1 (Pneumonia), which ensures that the model pays more attention to the underrepresented class during training.

\section{Results}
Through extensive experimentation, trials and testing, we developed a lightweight CNN model optimized for the efficient detection of Pneumonia using Chest X-Ray images. Our iterative approach involved testing various preprocessing and augmentation techniques, including varying input sizes (128x128, 224x224, 256x256) with both RGB and grayscale inputs. We carefully fine-tuned various hyperparameters, including the learning rate, class weights, the number of convolutional layers, filters and kernel sizes. Additionally, we tested our model on different pooling types (Max Pooling, Average Pooling), activation functions (such as ReLU and ELU), dropout layers with different rates, and different numbers of dense layers with their neurons. This comprehensive experimentation enabled us to determine the most effective model configuration, ensuring efficient and accurate pneumonia detection.

\subsection{Performance}
Our model demonstrated exceptional performance with an accuracy of 0.942, achieving a sensitivity/recall of 0.99. This indicates that our model effectively identified 99\% of the actual pneumonia cases in the test dataset. The precision of the  model was 0.92, with an F1-Score of 0.96. These results portray the model’s ability to detect pneumonia while maintaining strong overall accuracy and minimizing false negatives. Notably, even though our model is lightweight, it delivers such impressive results, showcasing its efficiency and effectiveness in pneumonia classification. 

\begin{figure}[h!]
    \centering
    \includegraphics[width=0.5\textwidth]{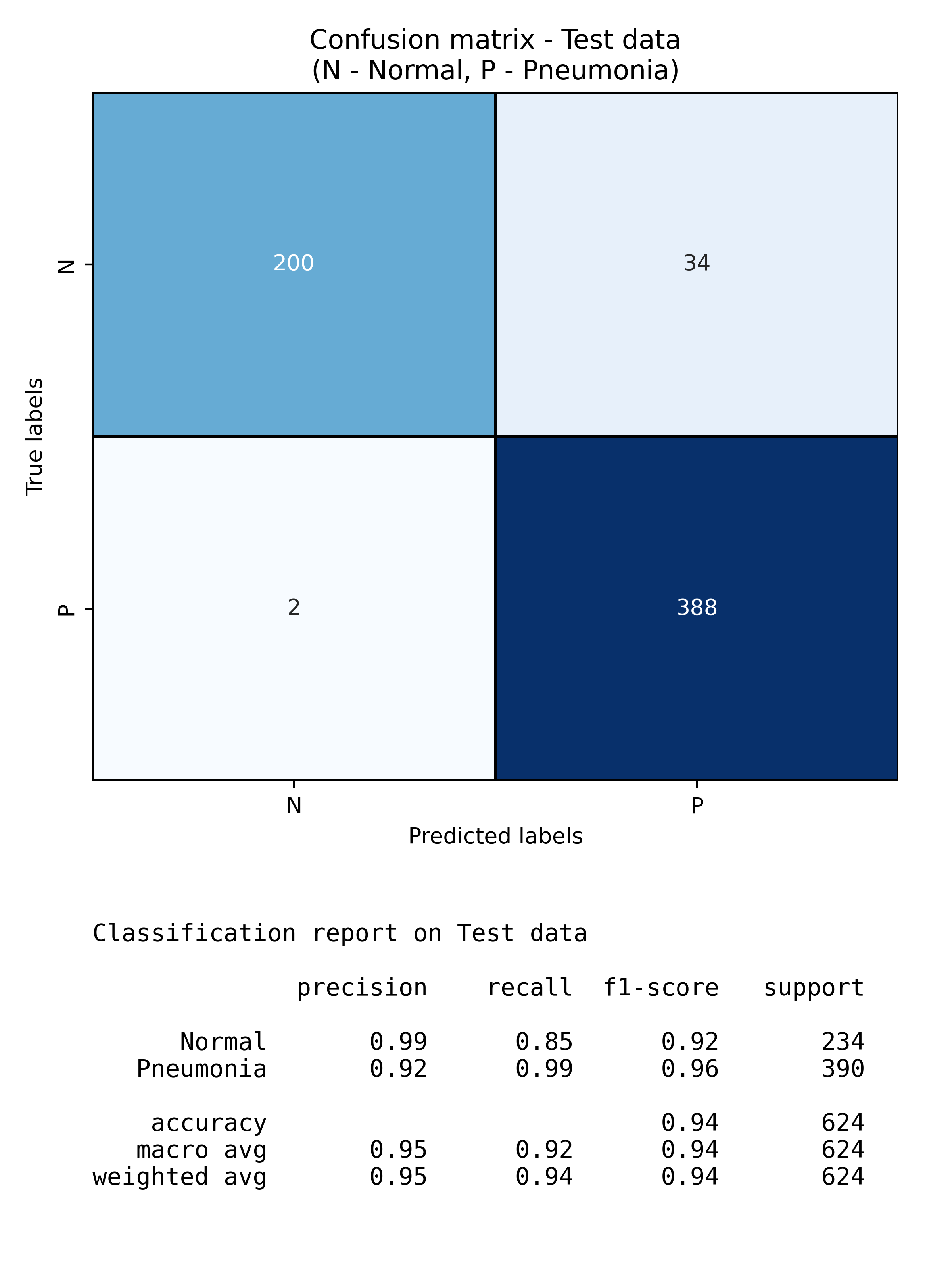} 
    \caption{Confusion matrix of the LightPneumoNet model on the independent test set.}
    \label{fig:fig5}
\end{figure}

\subsection{Comparison}
Our lightweight CNN model, designed for pneumonia detection, was compared with several existing models with respect to accuracy, precision, F1-score and most importantly recall (sensitivity). While most of the existing models rely on deeper architectures having numerous parameters, our model achieved similar or better performance with significantly fewer parameters, making it more efficient for real-world deployment. Furthermore, with a recall of 0.99, our model achieved outstanding results in detecting positive (pneumonia) cases, reducing the risk of missing pneumonia cases, which is critical in clinical settings.

\begin{table}[htbp]
    \centering 
    
    \caption{Detailed Evaluation and Comparison of existing Approaches with our Model}
    \label{tab:model_comparison} 
\resizebox{\textwidth}{!}{%
\begin{tabular}{llllllll}
\hline
\textbf{Ref.} & \textbf{Proposed Model} & \textbf{Accuracy} & \textbf{Precision} & \textbf{Recall} & \textbf{F1-Score} & \textbf{Considerations} &  \\ \hline
 & \textbf{\begin{tabular}[c]{@{}l@{}}LightPneumoNet (Our Model)\\ Dataset : {~\cite{Kermany2018}}\end{tabular}} & \textit{94.23\%} & \textit{91.94\%} & \textit{99.49\%} & \textit{95.57\%} & \textit{Lower Precision} &  \\ \hline
{~\cite{Racic2021}} & \begin{tabular}[c]{@{}l@{}}CNN (5 Convolutional Blocks)\\ Dataset : {~\cite{Kermany2018}}\end{tabular} & 88.9\% & 94.88\% & 87.6\% & 91.09\% & \begin{tabular}[c]{@{}l@{}}Lower Recall,\\ 8 million parameters\end{tabular} &  \\ \hline
{~\cite{Szepesi2022}} & \begin{tabular}[c]{@{}l@{}}CNN (10 Convolutional Blocks)\\ Dataset : {~\cite{Kermany2018}}\end{tabular} & (97.21±1.13)\% & (97.40±1.21)\% & (97.34±1.56)\% & (97.37±1.32)\% & 10 million parameters &  \\ \hline
{~\cite{Liang2020}} & \begin{tabular}[c]{@{}l@{}}CNN (49 Convolutional Layers)\\ Dataset : {~\cite{Kermany2018}}\end{tabular} & 90.5\% & 89.1\% & 96.7\% & 92.74\% & Lower Recall and Accuracy &  \\ \hline
{~\cite{Elshennawy2020}} & \begin{tabular}[c]{@{}l@{}}1) CNN\\ 2) LSTM-CNN\\ 3) ResNet152V2\\ 4) MobileNetV2\\ Dataset : {~\cite{Kermany2018}}\end{tabular} & \begin{tabular}[c]{@{}l@{}}1) 92.19\%           \\ 2) 91.80\%\\ 3) 99.22\%\\ 4) 96.48\%\end{tabular} & \begin{tabular}[c]{@{}l@{}}1) 95.57\%   \\ 2) 93.24\%\\ 3) 99.44\%\\ 4) 95.68\%\end{tabular} & \begin{tabular}[c]{@{}l@{}}1) 92.07\%\\ 2) 92.62\%\\ 3) 99.43\%\\ 4) 99.44\%\end{tabular} & \begin{tabular}[c]{@{}l@{}}1) 93.79\%\\ 2) 92.29\%  \\ 3) 99.44\%  \\ 4) 97.52\%\end{tabular} & \begin{tabular}[c]{@{}l@{}}Complex architecture,\\ Computationally Expensive\end{tabular} &  \\ \hline
{~\cite{Chhikara2020}} & \begin{tabular}[c]{@{}l@{}}InceptionV3\\ Dataset : {~\cite{Kermany2018}}\end{tabular} & 90.1\% & 90.71\% & 95.7\% & 93.1\% & Lower Recall and Accuracy &  \\ \hline
{~\cite{ElZein2021}} & \begin{tabular}[c]{@{}l@{}}Hybrid Model \\ (EfficientNetB0 as a transfer \\ learning-based model \& SVM \\ with hinge loss)\\ Dataset : {~\cite{Kermany2018}}\end{tabular} & 97.0\% & 100\% & 95.8\% & 97.9\% & \begin{tabular}[c]{@{}l@{}}Complex architecture,\\ Lower Recall\end{tabular} &  \\ \hline
{~\cite{Venu2020}} & \begin{tabular}[c]{@{}l@{}}Ensemble Model \\ (InceptionResNet, MobileNetV2, \\ Xception, DenseNet201, and \\ ResNet152V2)\\ Dataset : {~\cite{Kermany2018}}\end{tabular} & 98.46\% & 98.38\% & 99.53\% & 98.96\% & \begin{tabular}[c]{@{}l@{}}Complex architecture,\\ Computationally Expensive\end{tabular} &  \\ \hline
{~\cite{Rahman2020}} & \begin{tabular}[c]{@{}l@{}}1) AlexNet\\ 2) ResNet18\\ 3) DenseNet201\\ 4) SqueezeNet\\ Dataset : {~\cite{Kermany2018}}\end{tabular} & \begin{tabular}[c]{@{}l@{}}1) 94.5\%\\ 2) 96.4\%\\ 3) 98\%\\ 4) 96.1\%\end{tabular} & \begin{tabular}[c]{@{}l@{}}1) 93.1\%\\ 2) 95.4\%       \\ 3) 97\%          \\ 4) 98.5\%\end{tabular} & \begin{tabular}[c]{@{}l@{}}1) 95.3\% \\ 2) 97\%\\ 3) 99\% \\ 4) 94\%\end{tabular} & \begin{tabular}[c]{@{}l@{}}1) 94.18\%\\ 2) 96.19\%\\ 3) 97.98\%\\ 4) 96.19\%\end{tabular} & Complex architecture &  \\ \hline
{~\cite{Kundu2021}} & \begin{tabular}[c]{@{}l@{}}Ensemble Model (GoogLeNet,\\ ResNet-18, and DenseNet-121)\\ Trained on 1) Kermany {~\cite{Kermany2018}} and \\ 2) RSNA {~\cite{Wang2017}} dataset.\end{tabular} & \begin{tabular}[c]{@{}l@{}}1) 98.1\% \\ 2) 86.95\%\end{tabular} & \begin{tabular}[c]{@{}l@{}}1) 98.82\%\\ 2) 86.89\%\end{tabular} & \begin{tabular}[c]{@{}l@{}}1) 98.80\%\\ 2) 87.02\%\end{tabular} & \begin{tabular}[c]{@{}l@{}}1) 98.79\%\\ 2) 86.95\%\end{tabular} & \begin{tabular}[c]{@{}l@{}}Less Recall considering\\ the model’s complexity\end{tabular} &  \\ \hline
{~\cite{Erdem2021}} & \begin{tabular}[c]{@{}l@{}}CNN (6 Convolutional Blocks and\\ 6 Separable Convolutional Blocks)\\ Dataset : {~\cite{Kermany2018}}\end{tabular} & 88.62\% & 86.16\% & 97.3\% & 91.45\% & \begin{tabular}[c]{@{}l@{}}24 Million Parameters\\ (Computationally Expensive)\end{tabular} &  \\ \hline
{~\cite{Bhatt2023}} & \begin{tabular}[c]{@{}l@{}}Ensemble CNN Model\\ Dataset : {~\cite{Kermany2018}}\end{tabular} & 84.12\% & 80.04\% & 99.23\% & 88.56\% & Lower Precision and Accuracy &  \\ \hline
{~\cite{Iparraguirre2022}} & \begin{tabular}[c]{@{}l@{}}1) VGG16\\ 2) VGG19\\ 3) ResNet50\\ 4) InceptionV3\\ Dataset : {~\cite{Kermany2018}}\end{tabular} & \begin{tabular}[c]{@{}l@{}}1) 62.5\%\\ 2) 63.1\%\\ 3) 68.6\%\\ 4) 72.9\%\end{tabular} & \_\_ & \begin{tabular}[c]{@{}l@{}}1) 88.9\%\\ 2) 88.7\%\\ 3) 95.3\%\\ 4) 93.7\%\end{tabular} & \begin{tabular}[c]{@{}l@{}}1) 73.4\%\\ 2) 73.8\%\\ 3) 79.8\%\\ 4) 82\%\end{tabular} & Lower Classification Metrics &  \\ \hline
{~\cite{Eid2021}} & \begin{tabular}[c]{@{}l@{}}ResNet-Based SVM\\ Dataset : {~\cite{Kermany2018}}\end{tabular} & 98.13\% & 100\% & 96.4029\% & 98.1685\% & \begin{tabular}[c]{@{}l@{}}Less Recall considering the \\ model’s complexity\end{tabular} &  \\ \hline
{~\cite{Li2019}} & \begin{tabular}[c]{@{}l@{}}Pnet (CNN Model)\\ Dataset : PneuX-rays\end{tabular} & 92.79\% & 89.68\% & 92.59\% & 91.11\% & Lower Recall and Accuracy &  \\ \hline
{~\cite{Rajasenbagam2021}} & \begin{tabular}[c]{@{}l@{}}CNN (VGG19Net Architecture)\\ Dataset : Chest X-ray 8\end{tabular} & 99.3\% & 100\% & 98.6\% & 99.29\% & Complex architecture &  \\ \hline
\end{tabular}%
}
\end{table}

\FloatBarrier 

\section{Conclusion}
In this study, we developed an efficient, lightweight CNN architecture from the ground up for pneumonia detection using chest X-ray images, distinguishing our approach from many existing methods that heavily rely on transfer learning. Despite being lightweight, our model was able to achieve a high accuracy of 94.2\% and exhibited outstanding performance with a recall of 0.99 for detecting pneumonia. These outcomes demonstrate the model’s effectiveness in pneumonia classification while maintaining overall accuracy, offering a promising alternative to more complex and resource intensive architectures.

While our evaluation is conducted on a relatively small dataset of 5,856 chest X-ray images (Kermany et al.)~\cite{Kermany2018}, this constraint actually highlights the strength of our approach. Many existing models on this same dataset either fail to achieve above 90\% recall/accuracy or require significantly more complex architectures with millions of parameters to reach comparable performance. Our model's ability to achieve 99\% recall with only 388K parameters on this challenging, limited dataset demonstrates its efficiency and robustness, making it particularly valuable for scenarios where both data and computational resources are constrained.

Our results show that lightweight models, when properly designed, can work just as well or better than larger, transfer learning based models. This makes our model a more efficient choice for medical diagnostics, especially for deployment in resource-constrained environments, such as rural areas, where it is difficult to afford to deploy large models.


\end{document}